\let\ORIlabel\label
\let\ORIrefstepcounter\refstepcounter

\AddToHook{package/hyperref/before}{%
  \let\label\ORIlabel
  \let\refstepcounter\ORIrefstepcounter
}

\documentclass[onefignum,onetabnum]{siamart220329}

\usepackage{matlab-prettifier}
\usepackage{comment}
\usepackage{soul}

\usepackage{lipsum}
\usepackage{amsfonts,amssymb}
\usepackage{graphicx}
\usepackage{epstopdf}
\usepackage{algorithmic}
\usepackage{xcolor}

\ifpdf
  \DeclareGraphicsExtensions{.eps,.pdf,.png,.jpg}
\else
  \DeclareGraphicsExtensions{.eps}
\fi

\hypersetup{colorlinks,linkcolor=[RGB]{0,103,165},citecolor=[RGB]{180,0,0}}

\def\u{\mathbf{u}}
\def\v{\mathbf{v}}
\def\G{\mathbf{G}}

\usepackage{amsopn}

\headers{Flow map learning of nonlocal PDEs}{Z.Xu, Y.Li, Y.Zhang, and D.Xiu}

\title{Modeling Unknown Nonlocal PDE Systems via Flow Map Learning}

\author{Zhongshu Xu\thanks{Department of Mathematics, The Ohio State University, Columbus, OH 43210, USA. Emails: \email{\{xu.4202, li.14438, xiu.16\}@osu.edu}}. Funding: This work was partially supported by AFOSR FA9550-24-1-0237.
\and Ying Li\footnotemark[1]
\and Yanzhi Zhang\thanks{Department of Mathematics and Statistics, Missouri University of Science and Technology, Rolla, MO 65409, USA. Email: \email{zhangyanz@mst.edu}}.
\and Dongbin Xiu\footnotemark[1]}

\begin{document}

\maketitle

\begin{abstract}
Nonlocal partial differential equations arise in many applications but are often difficult to model and learn because of the presence of nonlocal operators. We present a flow-map learning (FML) framework for modeling unknown nonlocal PDEs directly from solution data. Rather than learning or approximating the underlying nonlocal operators, the proposed approach learns the finite-time evolution operator in either modal or nodal space. Two complementary formulations are developed for spectral and grid-based solution representations. Numerical experiments on one- and two-dimensional fractional diffusion and wave equations demonstrate accurate and stable long-time prediction using only short observation windows. The proposed approach provides an effective data-driven framework for learning unknown nonlocal dynamics without explicit evaluation of nonlocal operators.
\end{abstract}



\section{Introduction}

Nonlocal partial differential equations (PDEs) have attracted considerable attention because of their ability to model long-range interactions and anomalous transport phenomena that cannot be adequately described by classical local PDEs. Representative applications include anomalous diffusion, peridynamics in solid mechanics, porous media flow, and other multiscale systems governed by nonlocal interactions \cite{du2012analysis, kirkpatrick2016, metzler2000random, benson2000application, silling2000reformulation, caffarelli2011nonlinear}. 
Despite substantial progress in the development of nonlocal mathematical models, deriving predictive governing equations from first principles remains difficult in many practical settings because of complex multiscale interactions, incomplete physical knowledge, or limited observational access. Moreover, even when the governing equations are known, the associated nonlocal operators often introduce considerable analytical and computational complexity \cite{delia2020numerical, Duo2018, Duo2019, Zhou2024}.
These challenges have motivated increasing interest in data-driven approaches for modeling nonlocal dynamics directly from observations.

Data-driven methods for PDEs have received considerable attention in recent years and pursue several distinct objectives,  One class seeks to discover the governing equations in explicit analytical form. Such methods typically employ sparse regression over a prescribed dictionary of candidate differential operators and identify the terms that best fit the available data \cite{Rudy2017, Schaeffer2017, Long2019}. A second class focuses on approximating PDE solutions when the governing equations are known. Physics-informed neural networks (PINNs) and related methods incorporate differential-equation residuals, boundary conditions, and other physical constraints into the training objective and have been applied to a broad range of forward and inverse PDE problems \cite{Raissi2019, Zhu2019, Sun2020, Geneva2020}. 

Extensions of physics-informed methods to fractional and nonlocal PDEs include fractional and nonlocal PINNs, high-dimensional variants, and peridynamic formulations \cite{Pang2019, Pang2020, guo2022monte, Haghighat2021}.
Other approaches reduce the burden of directly evaluating nonlocal residuals through variational formulations, auxiliary extensions, or alternative numerical representations of known fractional operators \cite{gu2022deep, hao2023neural}. 
Neural-operator approaches have also been developed for learning nonlocal constitutive relations from paired field data, often with the nonlocal structure incorporated into the operator architecture \cite{jafarzadeh2024peridynamic}. Nevertheless, these methods generally require some representation of the underlying nonlocal mechanism during training, either through an explicit residual, a known reformulation, a prescribed architecture, or paired input-output data generated by the governing system. Since automatic differentiation does not directly apply to fractional and nonlocal operators \cite{lischke2020fractional, Nezza2012}, the training procedure may still require repeated evaluation or approximation of singular kernels, integral operators, or auxiliary discretizations. 

In this work, we pursue a different strategy based on Flow Map Learning (FML) \cite{ChurchillXiu_FML2023}.
Rather than identifying the governing equation or approximating its nonlocal operators, FML learns the finite-time evolution operator, or flow map, directly from solution data. Given two solution states separated by a fixed time interval, the method constructs a data-driven mapping between them. Once trained, the learned flow map can be applied recursively to generate long-time predictions. FML has been developed for autonomous and non-autonomous dynamical systems
\cite{qin2019data, Qin2021a}, unknown PDEs \cite{WU2020109307, CHEN2022110782, churchill2023dnn},
stochastic systems \cite{chen2024learning, xu2024modeling}, 
partially observed systems \cite{Fu2020, churchill2023dnn}, 
and fluid simulation and active control \cite{chen2026targeted, liu2026numerical}.

The objective of this paper is to extend FML to unknown nonlocal PDE systems and to investigate its effectiveness for long-time prediction. We develop two complementary formulations. The first operates in modal space, where the solution is represented through coefficients in a finite-dimensional basis. This formulation is particularly natural for periodic problems involving fractional operators, since the fractional Laplacian admits a simple representation in Fourier space. The second formulation operates directly in nodal space and learns the evolution of solution values on a spatial grid, making it applicable when a convenient modal representation is unavailable. In both cases, the learned FML model uses only solution snapshots and does not require explicit knowledge of the governing nonlocal operator.

The proposed methods are evaluated on one- and two-dimensional nonlocal diffusion and wave equations. The numerical results demonstrate that both modal- and nodal-space FML can provide accurate and stable long-time predictions from short training trajectories. Additional parameter-recovery experiments show that the predicted modal dynamics retain sufficient information to recover the underlying fractional orders over appropriate time intervals. These results suggest that FML provides a practical data-driven framework for modeling unknown nonlocal PDE systems without explicit approximation of their nonlocal operators.

The remainder of the paper is organized as follows. Section \ref{sec:setup} introduces
the problem setup and the observational data used for training. Section \ref{sec:modal}
presents the modal-space FML formulation, and
Section \ref{sec:nodal} develops the nodal-space FML formulation. Section \ref{sec:examples} contains numerical experiments for representative nonlocal diffusion and wave equations. Conclusions are presented in Section \ref{sec:conclusion}.

\section{Problem Setup and Data} 
\label{sec:setup}

We consider an autonomous nonlocal partial differential equation of the form
\begin{equation}
\label{pde}
\frac{\partial u}{\partial t}
=
\mathcal{L}(u),
\qquad
(x,t)\in \Omega\times \mathbb{R}^{+},
\end{equation}
subject to appropriate boundary and initial conditions
\begin{equation*}
u(x,0)=u_0(x),
\qquad x\in \Omega.
\end{equation*}
Here $\Omega\subset \mathbb{R}^{d}$, for $d=1,2,3$, is the spatial domain, $u=u(x,t)$ is the unknown solution, and $\mathcal{L}$ denotes a possibly nonlinear nonlocal operator. Throughout this paper, the explicit form of $\mathcal{L}$ is assumed to be unknown. 

\subsection{Data}

We shall assume that solution data are available at discrete time instances over a constant time step $\Delta t >0$, 
\begin{equation} \label{t_grid}
0=t_0<t_1<\cdots ,
\qquad
t_{\ell+1}-t_\ell=\Delta t,
\end{equation}
and a set of spatial grid points
\begin{equation} \label{x_grid}
X_N=\{x_1,\ldots,x_N\}\subset \Omega.
\end{equation}
We use vector notation to represent the solution 
\begin{equation} \label{u}
\u(t) = u(x,t)|_{X_N}= \left(u(x_1,t),\ldots,u(x_N,t)\right)^T
\in \mathbb{R}^{N}.
\end{equation}
The available solution data consist of $N_{\rm traj}\geq 1$ solution trajectories
\begin{equation} \label{trajectory_data}
\u^{(m)}(t_\ell),
\qquad
\ell=0,\ldots,L,
\qquad
m=1,\ldots,N_{\rm traj},
\end{equation}
generated from different initial conditions. Here $L\geq 1$ is the length of each trajectory and assumed to be a constant for notational convenience. The collection of these trajectory data is our raw data set
\begin{equation} \label{data_set}
    \mathcal{S} = \left\{
\u^{(m)}_0,\u^{(m)}_1,\dots,\u^{(m)}_{L}\right\}_{m=1}^{N_{\rm traj}},
\end{equation}
where the subscript $0\leq \ell\leq L$ provides the time index. Consequently, the time stamps $t_\ell$ are not required.

\subsection{Learning Objective}

For the unknown nonlinear problem \eqref{pde},
our objective is not to recover or approximate $\mathcal{L}$ itself, but rather to
learn the evolution of the solution from the observational data \eqref{data_set}.

Let $E_{\Delta t}$ denote the exact flow map of \eqref{pde} over a
time interval $\Delta t>0$:
\begin{equation}
\label{exact_flow}
u(\cdot,t+\Delta t) = E_{\Delta t} \left(u(\cdot,t)\right).
\end{equation}
FML seeks an approximation of this evolution operator directly
from data. Once such an approximation is learned, long-time prediction can be
performed by recursively applying the learned flow map operator, without requiring knowledge of the governing equation \eqref{pde}.

In the following two sections, we present two FML formulations, one in modal space and the other in nodal space. The modal formulation naturally aligns with the Fourier representation of fractional operators. 
The nodal formulation works directly with solution values on spatial grids.

\section{Learning Nonlocal PDEs in Modal Space}
\label{sec:modal}

\subsection{Motivation}

The fractional Laplacian plays a significant role in modeling anomalous diffusion and has become a fundamental nonlocal operator for describing long-range interactions in a wide range of scientific applications. 
It admits the following pseudo-differential representation \cite{kilbas1993fractional,Nezza2012}:
\begin{equation}
\label{eq:fractional_laplacian_fourier}
(-\Delta)^{\alpha/2}u
=
\mathcal{F}^{-1}
\left[
|\xi|^{\alpha}\mathcal{F}[u]
\right],\qquad \alpha > 0,
\end{equation}
where $\mathcal{F}$ denotes the Fourier transform and $\xi$ is the frequency
variable. This representation reveals an important feature of the fractional
Laplacian: Although it is nonlocal in physical space, it becomes a
multiplication operator in frequency space. Modal coefficients therefore
provide a natural representation for the dynamics of many nonlocal PDEs.

To further illustrate this, consider the fractional diffusion equation 
\begin{equation}
\label{eq:fractional_diffusion_example}
\partial_t u(x, t) = -(-\Delta)^{\alpha/2}u,\qquad x\in {\mathbb R}^d.
\end{equation}
Expanding the solution in Fourier modes
\begin{equation}
\label{eq:fourier_expansion}
u(x,t)
=
\sum_{k\in\mathbb{Z}^d}
\widehat{u}_k(t)e^{\mathrm{i}k \cdot x}
\end{equation}
with $\mathrm{i} = \sqrt{-1}$ being the imaginary unit and using the definition  \eqref{eq:fractional_laplacian_fourier}, we obtain
\begin{equation}
\label{eq:modal_ode_fractional_diffusion}
\frac{d\widehat{u}_k(t)}{dt}
=
-|k|^{\alpha}\widehat{u}_k, \qquad k \in {\mathbb Z}^d.
\end{equation}
Thus, in modal coordinates, the PDE is transformed into a dynamical system for the Fourier coefficients, analogous to the modal evolution obtained for classical local PDEs. In fact, when $\alpha=2$, the definition~\eqref{eq:fractional_laplacian_fourier} reduces to the spectral representation of the classical negative Laplacian and \eqref{eq:fractional_diffusion_example} corresponds to the standard diffusion equation. The exact one-step evolution is
\begin{equation}
\label{eq:exact_modal_flow_diffusion}
\widehat{u}_k(t+\Delta t)
=
\exp\left(-|k|^{\alpha}\Delta t\right)\widehat{u}_k(t), \qquad \alpha > 0.
\end{equation}
This simple example demonstrates that, despite its nonlocal nature in physical space, modal-space learning can still be natural for
nonlocal diffusion-type problems. For more general nonlocal PDEs, the modal
dynamics can be far more complex and unknown, but the evolution map can
still be learned from data.

\subsection{Finite-dimensional modal representation}

Let $V$ be a function space containing the solution of
\eqref{pde}, and let
\begin{equation}
\label{eq:finite_space}
V_n
=
{\rm span}\{\phi_1,\ldots,\phi_n\}
\subset V
\end{equation}
be an $n$-dimensional approximation space, where 
$\{\phi_j\}_{j=1}^{n}$ are the basis functions. 
We approximate the solution $u\in V$ by $u_n\in V_n$ through a projection operator $P_n:V\rightarrow V_n$. The approximate solution $u_n$ can be expressed as
\begin{equation}
\label{modal_approximation}
u_n(x,t) = P_n u(x,t) = \sum_{j=1}^{n} v_j(t)\phi_j(x),
\end{equation}
where
\begin{equation}
\label{eq:modal_vector}
\v(t) = (v_1(t),\ldots,v_n(t))^T
\in \mathbb{R}^{n}
\end{equation}
is the vector of modal coefficients. The modal coefficients depend on the choice of the approximation space $V_n$ and the projection operator $P_n$.

Define the coefficient-to-function map
\begin{equation}
\label{eq:Pi_map}
\Pi:\mathbb{R}^{n}\rightarrow V_n,
\qquad
\Pi \v
=
\sum_{j=1}^{n} v_j\phi_j .
\end{equation}
When the basis functions are linearly independent, $\Pi$ is a one-to-one
correspondence between coefficient vectors in $\mathbb{R}^{n}$ and functions in
$V_n$.

Consider the infinite-dimensional
flow map $E_{\Delta t}$ \eqref{exact_flow}:
\begin{equation}
    E_{\Delta t}: V\to V, \qquad u(\cdot, t+\Delta t) = E_{\Delta t} (u(\cdot,t)).
\end{equation}
Upon projecting the solutions via \eqref{modal_approximation} onto the finite-dimensional space $V_n$, i.e.,
\begin{equation} \label{M:proj}
    u_n(\cdot,t) = P_n u(\cdot,t), \qquad u_n(\cdot, t+\Delta t) = P_n u(\cdot,t+\Delta t),
\end{equation}
one can define a finite dimensional evolution operator $\widetilde{E}_{\Delta t}$:
\begin{equation} \label{M:map}
    \widetilde{E}_{\Delta t}: V_n\to V_n, \qquad 
    u_n(\cdot, t+\Delta t) = \widetilde{E}_{\Delta t} (u_n(\cdot,t)).
\end{equation}
Subsequently, in modal
coordinates, this finite-dimensional evolution is represented by a map
\begin{equation} \label{M:def}
M_{\Delta t}:\mathbb{R}^{n}\rightarrow\mathbb{R}^{n}, \qquad
\v(t+\Delta t) = M_{\Delta t}(\v(t)),
\end{equation}
which satisfies
\begin{equation} \label{M}
    M_{\Delta t}=\Pi^{-1} \widetilde{E}_{\Delta t} \Pi.
\end{equation}

\subsection{Training of modal-space FML}

The purpose of modal-space FML is to approximate $M_{\Delta t}$ \eqref{M} directly from the data \eqref{data_set}. The derivation \eqref{M:proj}--\eqref{M:def} readily gives us the following learning procedure.

\subsubsection{Training Data Set}

Given the training data set of the solution snapshots $\u$
\eqref{data_set}, our first task is to convert it into a training data set for the modal coefficients $\v$. This is accomplished by projecting each snapshot $\u$ onto $V_n$ via the projection $P_n$ \eqref{modal_approximation}. However, since $\u\in\mathbb{R}^N$ is finite dimensional, we need to define a discrete projection
$\hat{P}_n:\mathbb{R}^N\to V_n$ such that
\begin{equation}
\label{discrete_proj}
\hat{u}_n(x,t) = \hat{P}_n \u(t) = \sum_{j=1}^{n} \hat{v}_j(t)\phi_j(x),
\end{equation}
where
\begin{equation}
\label{vhat}
\hat{\v}(t) = (\hat{v}_1(t),\ldots,\hat{v}_n(t))^T
\in \mathbb{R}^{n}
\end{equation}
is the vector of modal coefficients. These coefficients are determined by the choice of the discrete projection operator $\hat{P}_n$, which is essentially a numerical approximation of $P_n$ via the finite grid $X_N$. The resulting difference between $u_n$ and $\hat{u}_n$, resp. between $\v$ and $\hat{\v}$, is the so-called aliasing error. Upon applying the discrete projection $\hat{P}_n$ to each solution snapshot in the training data set \eqref{data_set} and then $\Pi^{-1}$ \eqref{eq:Pi_map} to retract the (approximate) modal coefficients, we obtain the following data set
\begin{equation}
\label{modal_data}
\mathcal{D} = \left\{
\hat{\v}^{(m)}_0,\hat{\v}^{(m)}_1,\ldots, \hat{\v}^{(m)}_L
\right\}_{m=1}^{N_{\rm traj}},
\end{equation}
which will be used as our training data set for model-space learning.

\subsubsection{FML Model Construction and Prediction}

In modal-space FML, we seek an operator $\widetilde{M}_{\Delta t}:\mathbb{R}^n\to\mathbb{R}^n$ as an approximation to the true unknown modal flow map $M_{\Delta t}$ \eqref{M:def}. More specifically, we use a residual neural network (ResNet) to represent it, i.e.,
\begin{equation}
\label{M_resnet}
\widetilde{M}_{\Delta t} = \mathbf{I}_n+\mathcal{N}(\cdot;\Theta),
\end{equation}
where $\mathbf{I}_n$ is the identity matrix of size $n\times n$ and $\mathcal{N}:\mathbb{R}^n\to\mathbb{R}^n$ is the mapping operator of a fully connected DNN with $n$ input nodes, $n$ output nodes, and hyperparameters $\Theta$. 

To learn this operator, we use the training data set \eqref{modal_data}. We randomly sample $N_{\rm pair}>1$ pairs of consecutive entries from \eqref{modal_data}
\begin{equation} \label{pair}
\left(\hat{\v}^{(m_k)}_{i_k},\hat{\v}^{(m_k)}_{i_k+1}\right), \qquad
k=1, \dots, N_{\rm pair}.
\end{equation}
Typically, such random sampling is conducted via uniform distribution over the subscript indices of \eqref{modal_data}. We then seek the operator \eqref{M_resnet} by minimizing the following mean-squared loss
\begin{equation} \label{Loss1}
    \mathcal{L}_1(\Theta) = \frac{1}{N_{\rm pair}}
\sum_{k=1}^{N_{\rm pair}}
\left\|
\widetilde{M}_{\Delta t}
\left(\hat{\v}^{(m_k)}_{i_k};\Theta\right)
-
\hat{\v}^{(m_k)}_{i_k+1}
\right\|_2^2.
\end{equation}
In practice, multi-step loss (or, roll-over loss) is preferred because it enhances the numerical stability and accuracy of the trained model (\cite{ChurchillXiu_FML2023}). Let $n_r\geq 1$ be the number of multi-step loss. We then randomly sample, from the data set \eqref{modal_data}, $N_{\rm seq}>1$ sequences of consecutive entries,
\begin{equation} \label{seq}
\left(\hat{\v}^{(m_k)}_{i_k},\hat{\v}^{(m_k)}_{i_k+1},\cdots, \hat{\v}^{(m_k)}_{i_k+n_r}\right), \qquad
k=1, \dots, N_{\rm seq}.
\end{equation}
The multi-step loss function is defined as
\begin{equation} \label{Loss_N}
    \mathcal{L}_{n_r}(\Theta) = \frac{1}{N_{\rm seq}}
\sum_{k=1}^{N_{\rm seq}}
\frac{1}{n_r} \sum_{j=0}^{n_r-1}
\left\|
\widetilde{M}_{\Delta t}
\left(\widetilde{\v}^{(m_k)}_{i_k+j};\Theta\right)
-
\hat{\v}^{(m_k)}_{i_k+j+1}
\right\|_2^2, 
\end{equation}
where
\begin{equation}
\widetilde{\v}^{(m_k)}_{i_k} = \hat{\v}^{(m_k)}_{i_k}, \qquad
\widetilde{M}_{\Delta t}
\left(\widetilde{\v}^{(m_k)}_{i_k+j};\Theta\right)
=\widetilde{\v}^{(m_k)}_{i_k+j+1}, \quad j=0,\dots,n_r-1.
\end{equation}
When $n_r=1$, this loss becomes the one-step loss \eqref{Loss1}.

Upon minimizing the loss to a satisfactorily low level, the FML model training is considered finished, and the hyperparameters $\Theta$ become fixed. We then obtain an FML model for system prediction. Given a new initial condition $\v_0$, the FML prediction follows
\begin{equation}  \label{modal_pred}
\left\{
\begin{aligned}
    & \widetilde{\v}_0 = \v_0, \\
    & \widetilde{\v}_{k+1}
=
\widetilde{M}_{\Delta t}(\widetilde{\v}_k), \quad 
\qquad k=0,1,\dots
\end{aligned}
\right.
\end{equation}
The solution at any time instance can be obtained by applying \eqref{eq:Pi_map},
i.e., $\widetilde{u}_n(x,t_k) = \Pi\widetilde{\v}_k$, $\forall k$.

\section{Learning Nonlocal PDEs in Nodal Space}
\label{sec:nodal}

\subsection{Motivation}

The modal-space formulation is natural when an appropriate basis is available 
and the solution data can be readily projected using the basis. 
In many applications with complex geometry, modal-space representation can be cumbersome, 
if not impossible. It is therefore useful to construct a learning framework directly in physical space. 

Consider the solution vector $\u(t)$ \eqref{u} over the spatial grid $X_N$ \eqref{x_grid}, its semi-discrete evolution follows
\begin{equation} \label{semi-discrete}
\frac{d \u(t)}{dt} = \mathbf{F}(\u),
\end{equation}
where $\mathbf{F}:\mathbb{R}^{N}\rightarrow \mathbb{R}^{N}$ is an unknown evolution operator 
containing discrete representations of various operators in the unknown governing PDE equation \eqref{pde}. 
For nonlocal PDEs involving nonlocal differential operators, their discretization can be expressed in terms of matrix-vector products. 
For example, the fractional Laplacian may be approximated in nodal space as \cite{Duo2018, Duo2019, Zhou2024}
\begin{equation}
(-\Delta)^{\alpha/2}u
\approx
\mathbf{D}_{\alpha}\u,
\end{equation}
where $\mathbf{D}_{\alpha}\in\mathbb{R}^{N\times N}$ is a dense differentiation matrix.

Integrating the semi-discrete system \eqref{semi-discrete} over one time step $[t_n, t_{n+1}]$ gives the exact identity
\begin{equation} \label{integral}
    \u(t_{n+1}) = \u(t_n) + \int_{t_n}^{t_{n+1}}
    \mathbf{F}(\u(t), \mathbf{D}_{\alpha_1}\u(t), \mathbf{D}_{\alpha_2}\u(t),\cdots)\, dt,
\end{equation}
where $\mathbf{D}_{\alpha_j}$, $j=1,\dots$, are the differentiation matrices for various differential operators.
The formulation is Markovian and effectively defines a mapping 
\begin{equation} \label{G}
  \mathbf{G}_{\Delta t}:\mathbb{R}^N\to\mathbb{R}^N, \qquad \u(t_{n+1}) = \mathbf{G}_{\Delta t}(\u(t_n)).
\end{equation}
We remark again that this is an exact formulation and not an Euler forward approximation of \eqref{semi-discrete}. 
Hereafter, we shall refer to $\mathbf{G}_{\Delta t}$ as the nodal flow map.

\subsection{Nodal-space FML model}

We approximate the nodal flow map $\mathbf{G}_{\Delta t}$ by a residual neural network
\begin{equation} \label{nodal_resnet}
\widetilde{\G}_{\Delta t} = \mathbf{I}_N +\mathcal{N}(\cdot;\Theta),
\end{equation}
where $\mathbf{I}_N$ is the identity matrix of size $N\times N$ and $\mathcal{N}:\mathbb{R}^{N}\rightarrow\mathbb{R}^{N}$ is the mapping operator of a DNN with $N$ input nodes, $N$ output nodes, and parameters $\Theta$.

\begin{figure}[!htbp]
  \centering
  \includegraphics[width=\textwidth]{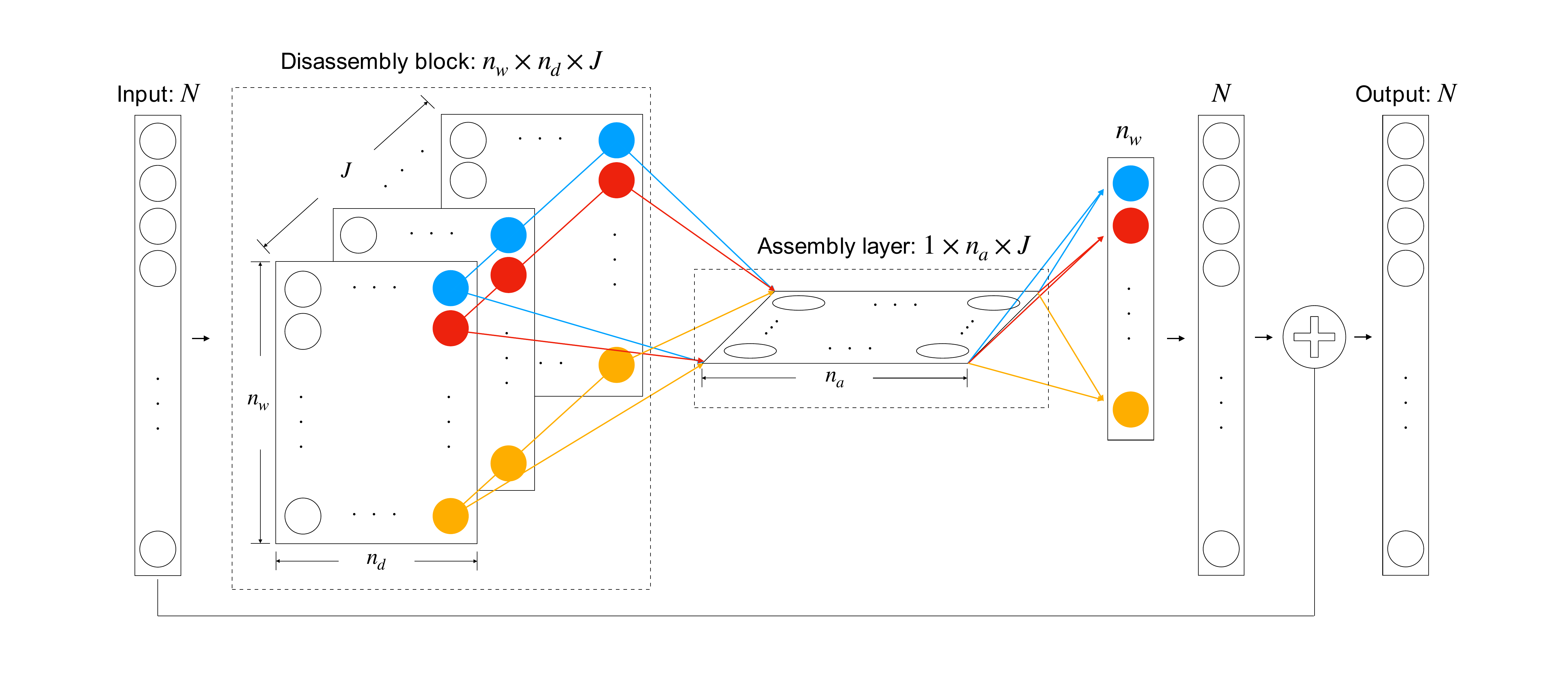}
  \caption{Structure of the disassembly--assembly network used for nodal-space FML.}
  \label{fig:disassembly_assembly}
\end{figure}

Following \cite{CHEN2022110782}, we take $\mathcal{N}$ in \eqref{nodal_resnet} to
be a disassembly--assembly network, whose structure is illustrated in
Figure~\ref{fig:disassembly_assembly}. It consists of a disassembly block, an
assembly layer, and an output layer.

The disassembly block comprises $J\geq 1$ fully connected feedforward channels
operating in parallel, each with $n_d$ hidden layers of $n_w$ neurons. Let
\begin{equation}
\mathcal{F}_i:\mathbb{R}^{N}\rightarrow\mathbb{R}^{n_w},
\qquad i=1,\ldots,J,
\end{equation}
denote the operator defined by the $i$th channel. 
Collecting the $J$ channels, the disassembly block defines the operator
\begin{equation}
\mathcal{D}=\bigotimes_{1\leq i\leq J}\mathcal{F}_i:
\mathbb{R}^{N}\rightarrow\mathbb{R}^{n_w\times J}.
\end{equation}
The assembly layer is a fully connected feedforward network with $n_a$ hidden layers of $J$ neurons. 
It operates componentwise on the output of the
disassembly block, taking a $J$-dimensional input to a scalar,
\begin{equation}
\mathcal{A}:\mathbb{R}^{J}\rightarrow\mathbb{R},
\end{equation}
so that the $n_w\times J$ output of the disassembly block is reduced to an
$n_w$-vector. The output layer
\begin{equation}
\mathcal{O}:\mathbb{R}^{n_w}\rightarrow\mathbb{R}^{N}
\end{equation}
is a fully connected feedforward network that maps the assembled $n_w$-vector
back to the $N$-dimensional nodal space. With these operators, the network in
\eqref{nodal_resnet} reads
\begin{equation}
\label{eq:nodal_network}
\widetilde{\G}_{\Delta t}
=\mathbf{I}_N+\mathcal{N}
=\mathbf{I}_N+\mathcal{O}\circ\mathcal{A}\circ\mathcal{D},
\end{equation}
where the residual term
$\mathcal{O}\circ\mathcal{A}\circ\mathcal{D}$ gives the one-step increment of
the nodal flow map. The construction of the disassembly-assembly structure is motivated by the representation of the various unknown differentiation matrices in \eqref{integral}. For the mathematical details, see \cite{CHEN2022110782}.

An essential feature of the proposed DNN is that there is no need to explicitly construct a fractional differentiation matrix or evaluate a singular integral operator. All the nonlocal dependence of the unknown system is embedded in the nodal flow map $\mathbf{G}_{\Delta t}$, to be learned from the data \eqref{data_set}.

\subsection{Training of nodal-space FML}

The training of the nodal FML operator \eqref{nodal_resnet} can be accomplished by directly using the training data set \eqref{data_set}. Again, let $n_r\geq 1$ be the number of multi-step loss. We randomly sample $N_{\rm seq}>1$ sequences of consecutive entries from \eqref{data_set},
\begin{equation} \label{u_seq}
\left({\u}^{(m_k)}_{i_k},{\u}^{(m_k)}_{i_k+1},\cdots, {\u}^{(m_k)}_{i_k+n_r}\right), \qquad
k=1, \dots, N_{\rm seq}.
\end{equation}
Let 
$$
\widetilde{\G}_{\Delta t}^\ell =\underbrace{\widetilde{\G}_{\Delta t} \circ \widetilde{\G}_{\Delta t} \circ \cdots \circ \widetilde{\G}_{\Delta t}}_{\ell \text{ times}}, \qquad \ell\geq 1,
$$
be $\ell$ times recurrent operations of $\widetilde{\G}_{\Delta t}$. We then seek to minimize the following loss 
\begin{equation}
\label{eq:nodal_recurrent_loss}
\mathcal{L}_{n_r}(\Theta)
=
\frac{1}{N_{\rm seq}}
\sum_{k=1}^{N_{\rm seq}}
\frac{1}{n_r}
\sum_{\ell=1}^{n_r}
\left\|
\widetilde{\G}_{\Delta t}^{\ell}
\left(\u^{(m_k)}_{i_k}; \Theta\right)
-
\u^{(m_k)}_{i_k+\ell}
\right\|_2^2.
\end{equation}
Upon satisfactory training, the parameters $\Theta$ are fixed, and we obtain a nodal FML predictive model. For any given solution snapshot $\u_0$ as the initial condition, the FML model prediction follows
\begin{equation} \label{nodal_pred}
\left\{
\begin{aligned}
    & \widetilde{\u}_0 = \u_0, \\
    & \widetilde{\u}_{k+1}
=
\widetilde{\mathbf{G}}_{\Delta t}(\widetilde{\u}_k), \quad 
\qquad k=0,1,\dots
\end{aligned}
\right.
\end{equation}

Compared with the methods that require evaluating nonlocal operators during training, the nodal-space FML formulation uses only solution snapshots.
This feature is particularly attractive for unknown nonlocal PDEs, where the operator form, the kernel, or the fractional order are often not available.
\section{Numerical Examples}
\label{sec:examples}

This section evaluates the nodal- and modal-space FML methods on three nonlocal PDEs. The governing
equations are used only to generate synthetic solution data, either analytically or with high-resolution
numerical solvers, for training and validation of the FML models. 
In all error plots, the errors are computed from the discrete solution vectors on the corresponding computational grid.
Unless otherwise specified, the reported relative error is defined as
$$
\frac{\|\u_{\rm pred}-\u_{\rm ref}\|_{2}}{\|\u_{\rm ref}\|_{2}},
$$
where $\u_{\rm pred}$ is the FML model prediction and $\u_{\rm ref}$ the reference solution, over the same grid.

\subsection{{Anomalous Diffusion Equation}}

We first consider the 1D  anomalous diffusion equation \cite{delia2020numerical}:
\begin{equation}\label{diff1D}
\left\{
\begin{aligned}
    &\frac{\partial u(x, t)}{\partial t}=-(-\Delta)^{\frac{\alpha}{2}} u(x, t),\quad  x \in (0,2\pi), \ \ t>0, \\
    &u(x, 0) = u_0(x),\quad  x \in [0,2\pi].
\end{aligned}
\right.
\end{equation}
with periodic boundary conditions. We set $\alpha=1.5$ in this experiment.

The training data are the numerical solution of \eqref{diff1D}, 
obtained using the Fourier pseudo-spectral method for spatial discretization and exact time integration \cite{Duo2016}, 
with $N_x = 51$ spatial points. 
We generate $N_{\rm train} = 10^5$ training sequences from random initial conditions of the form
\begin{equation}
u(x, 0)=a_0+\sum_{n=1}^{N_c}\big(a_n \cos (n x)+b_n \sin (n x)\big),  \quad  x \in [0,2\pi],
\label{eq: diffu1d IC}  
\end{equation}
We sample $N_c \sim U\{0,1, \ldots, 7\}$, $a_0 \sim U[-2,2]$, and
$a_n, b_n \sim U[-1/n,1/n]$ for $1 \leq n \leq N_c$, where $U$ denotes the uniform distribution.
For each initial condition, we evolve \eqref{diff1D} for 30 steps with $\Delta t = 0.05$ (up to $T=1.5$).
We then select six consecutive snapshots, corresponding to five recurrent prediction steps, to form
$$
\left\{\u^{(i)}(t_i),\u^{(i)}(t_i+\Delta t),\ldots,
\u^{(i)}(t_i+5\Delta t)\right\}_{i=1}^{N_{\rm train}},
$$
where $t_i \sim U\{0,\Delta t,\ldots,25\Delta t\}$. Thus, $N_{\rm train}$ denotes the number
of training sequences rather than the number of snapshots.
For nodal FML, we use a disassembly--assembly network with $J=3$, $n_d=n_a=1$, and $n_w=51$.
For modal FML, we use a one-block ResNet with six hidden layers of 50 neurons each.
Both models use a five-step recurrent loss and a cyclic learning rate that ranges from $10^{-7}$ to
$10^{-3}$ with exponential decay factor $0.9999997$. We train each network for 10,000 epochs with a batch size of 50.

\begin{figure}[htbp!]
	\centering
	\includegraphics[width=0.8\textwidth]        
        {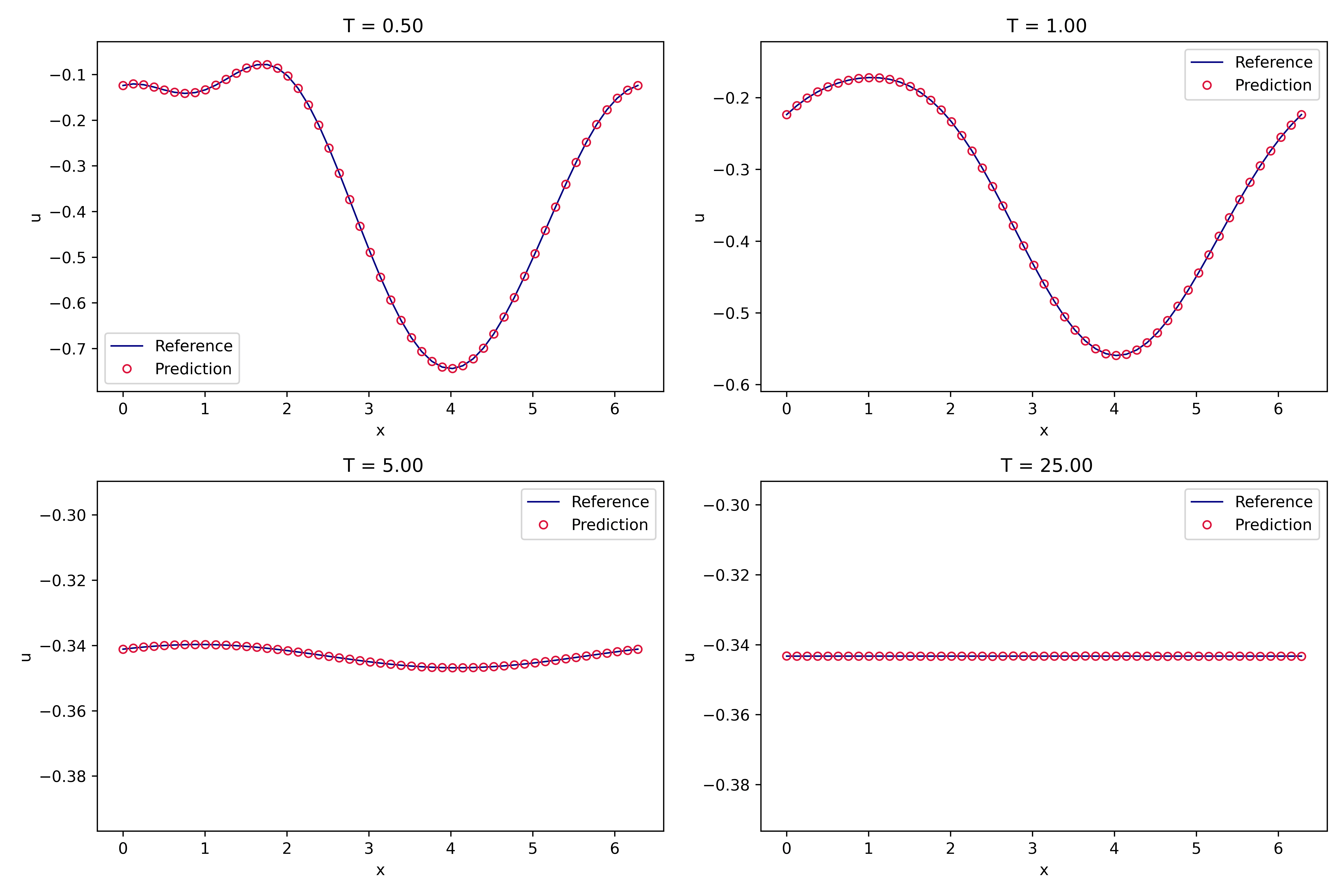}
	\caption{Reference and nodal FML predictions for the solution of \eqref{diff1D}.}
	\label{fig: dif1d traj}
\end{figure}

\begin{figure}[htbp]
	\centering
	\includegraphics[width=0.8\textwidth]{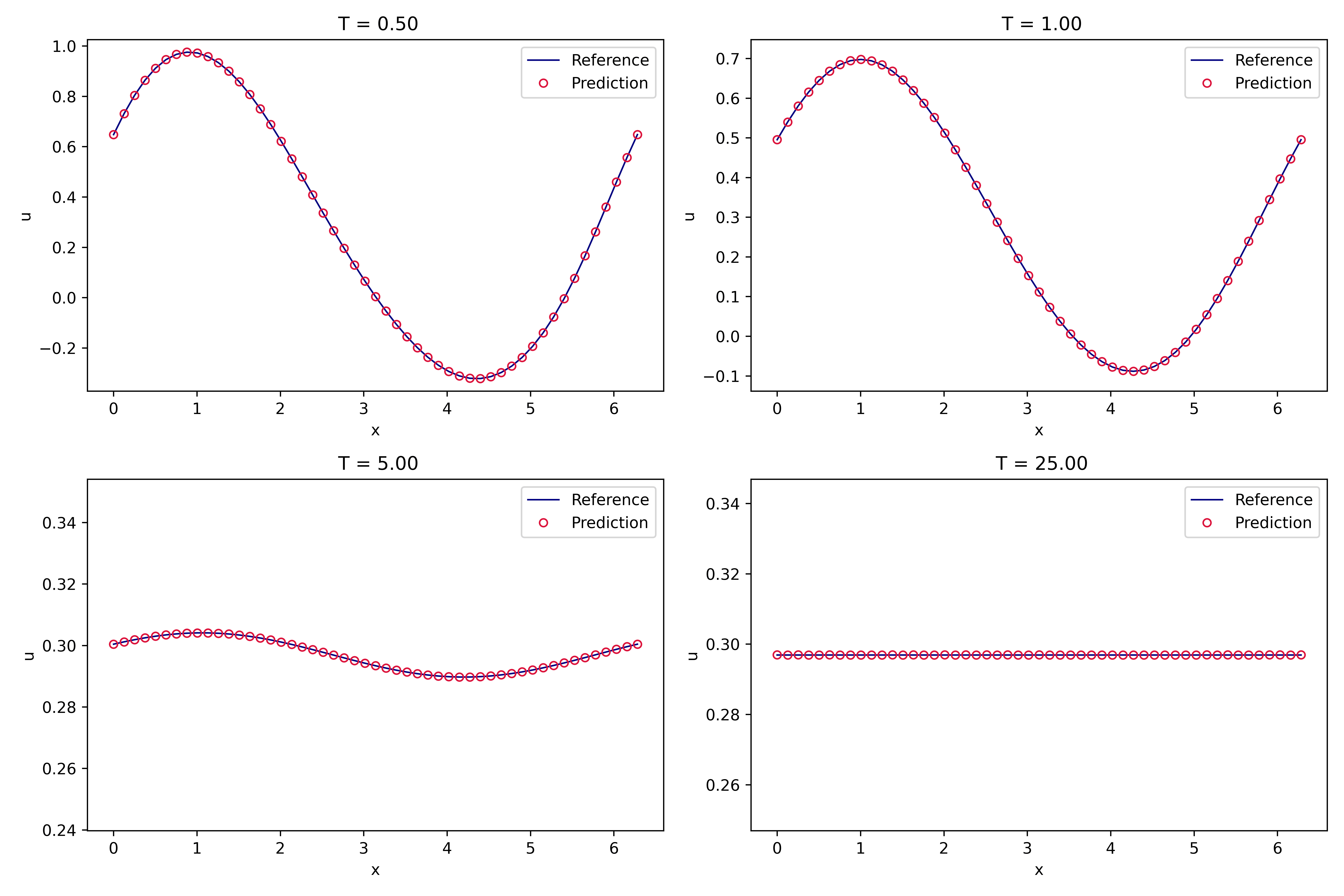}
	\caption{Reference and modal FML predictions for the solution of  \eqref{diff1D}.}
	\label{fig: diff1d modal traj}
\end{figure}

\begin{figure}[htbp!]
    \centering
    (a)\includegraphics[width=0.4\textwidth]{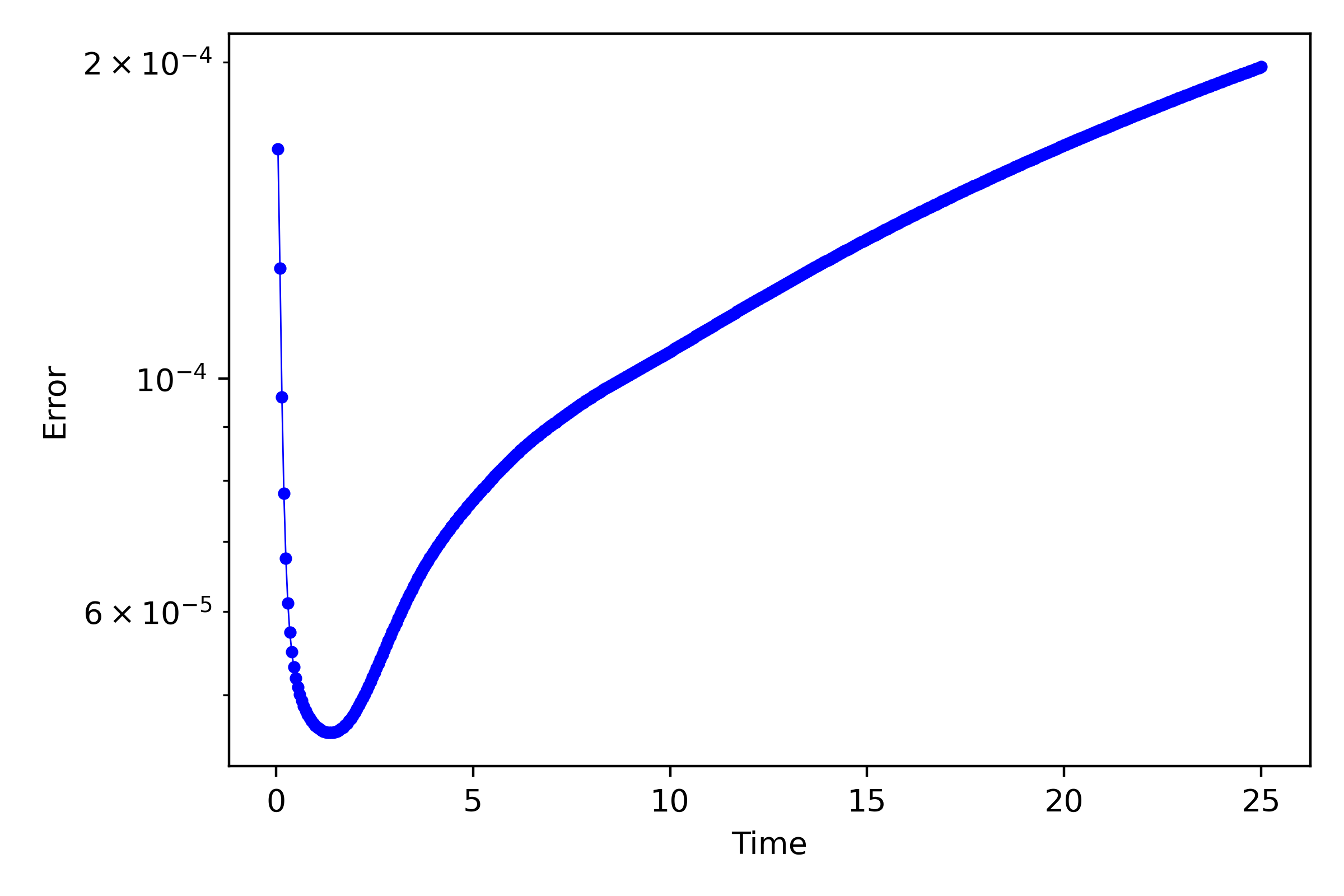}
    (b)\includegraphics[width = 0.4\textwidth]{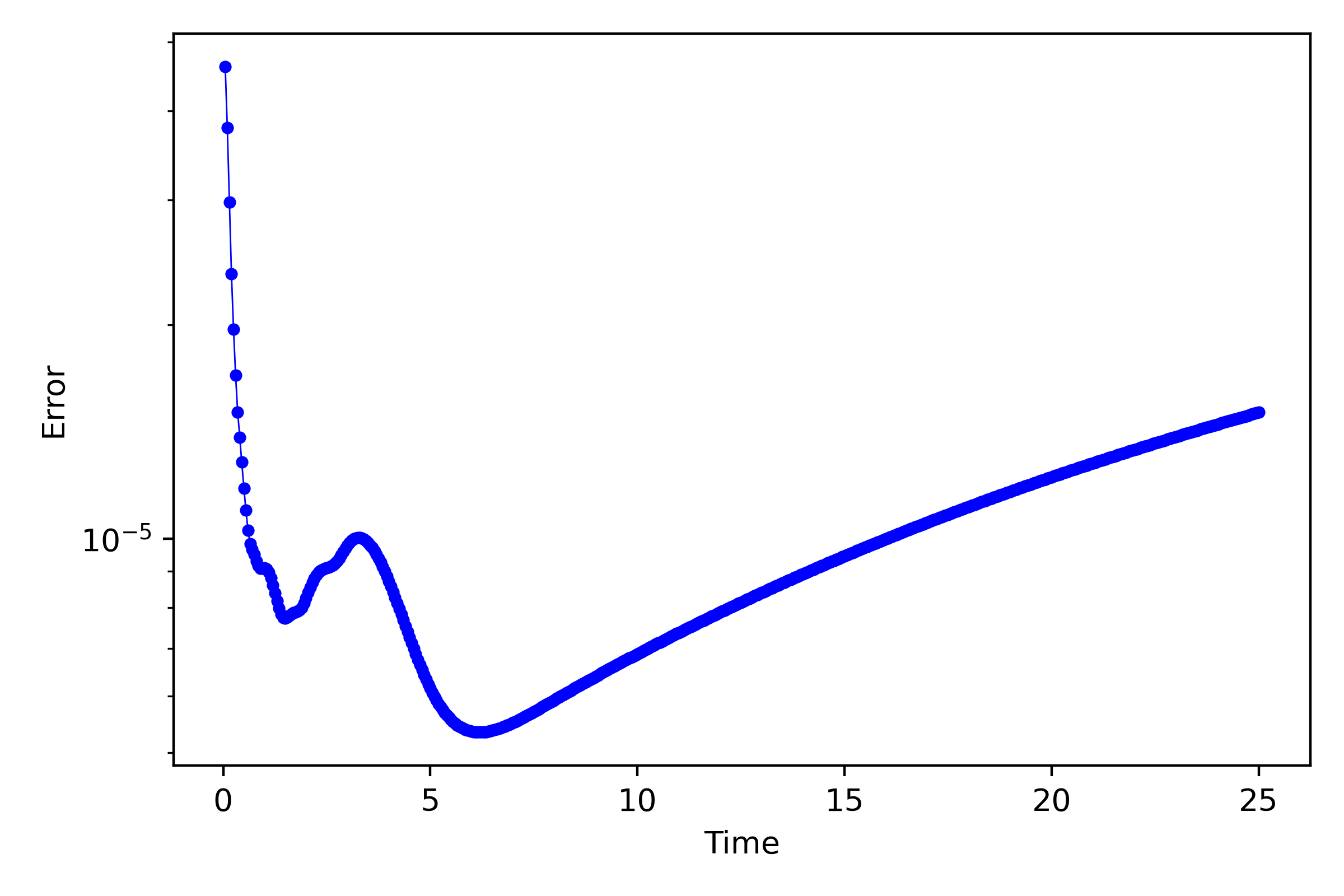}
    \caption{Average $\ell_2$ error over 100 test samples from nodal (a) and modal (b) FML.}
    \label{fig: diff1d error}
\end{figure}

For validation, we generate 100 test samples independently from the same distribution and evaluate 10
independently trained models over 500 time steps (up to $T=25$). Figures~\ref{fig: dif1d traj}
and \ref{fig: diff1d modal traj} show nodal and modal predictions, respectively, for one test sample.
Both predictions follow the reference trajectory up to $T=25$. Figure~\ref{fig: diff1d error} shows
the mean $\ell_2$ errors over the 100 test samples: approximately $10^{-4}$ for nodal FML and
$10^{-5}$ for modal FML.
This demonstrates that the learned FML model remains accurate under recursive application for several hundred prediction steps, despite being trained using only short trajectory segments.

\subsection{Nonlocal Wave Equation} 
We consider the 1D nonlocal wave equation of the form:
\begin{equation}\label{wave}
\left\{
\begin{aligned}
	&  u_{tt} (x, t)=-D(-\Delta)^{\frac{\alpha}{2}} u(x, t), \quad \ \text {for} \ \, x\in \Omega, \ \  t>0, \\
	& u(x, 0)=u_0(x), \quad  u_t(x, 0)=v_0(x), \quad \  \text {for} \ \, x\in \bar{\Omega}
\end{aligned}
\right.
\end{equation}
with periodic boundary conditions. 
In this experiment, $\Omega=(0,2\pi)$, $\alpha = 0.5$, and $D=1$.
The numerical solutions are generated with the Fourier pseudo-spectral method in space with $N_x = 51$ and exact time integration.

For the training set, we generate $N_{\rm train}=10^6$ sequences from random initial displacement
$u_0$ and velocity $v_0$. Their nodal vectors are sampled from periodic Gaussian vectors as follows.
Set the grid points to be $\theta_j = \frac{2\pi j}{N_x-1}$ for $j=0, \ldots, N_x-1$,
where the periodic grid includes the duplicated endpoint.
We generate $\mathbf{s} = (s_0,\dots,s_{N_x-1})$ such that
$\operatorname{Cov}(s_j,s_k)=\exp \big(-\left(\theta_j-\theta_k\right)^2 / b^2\big)$ with $b$ the correlation length by
\begin{align*}
s_j &=\sum_{n=-K_G}^{K_G} R_n e^{-\mathrm{i} n \theta_j},\quad j=0,\ldots,N_x-1,\\
R_n&\sim \mathcal{N}(0, C_n),\qquad
C_n = \int_{0}^{2\pi} e^{-\frac{\theta^2}{b^2}} \cos(n\theta) d \theta.
\end{align*}
The sampled displacement and velocity vectors are $\u_0=\mathbf{s}+m_u$ and
$\mathbf{v}_0=\mathbf{s}+m_v$, where $m_u\sim U[-1,1]$, $m_v\sim U[-0.1,0.1]$,
$b\sim U[0.1,1.1]$, and $K_G=10$.
We represent the discrete wave state by
$$
\mathbf{z}(t)=
\begin{pmatrix}
\u(t)\\
\dot{\u}(t)
\end{pmatrix}
\in\mathbb{R}^{2N_x}.
$$
For each initial condition, we solve \eqref{wave} for 20 steps with $\Delta t=0.05$ and form
one-step training pairs for nodal FML:
$$
\Big\{\mathbf{z}^{(i)}(t_i),\mathbf{z}^{(i)}(t_i+\Delta t)\Big\}_{i=1}^{N_{\rm train}},
$$
where $t_i\sim U\{0,\Delta t,\ldots,19\Delta t\}$.

For nodal FML, we use a disassembly--assembly network with input dimension $N=2N_x=102$,
$J=3$, $n_d=n_a=1$, and $n_w=51$.
For modal FML, we use the same $N_{\rm train}=10^6$ trajectories and select six consecutive
snapshots, corresponding to five recurrent prediction steps:
$$
\left\{\mathbf{z}^{(i)}(t_i),\mathbf{z}^{(i)}(t_i+\Delta t),\ldots,
\mathbf{z}^{(i)}(t_i+5\Delta t)\right\}_{i=1}^{N_{\rm train}},
$$
where $t_i\sim U\{0,\Delta t,\ldots,15\Delta t\}$. The modal model is a two-block
ResNet with three hidden layers per block and 40 neurons per layer, trained with a five-step recurrent loss.

\begin{figure}[htbp]
	\centering
	(a)\includegraphics[width=0.4\textwidth]{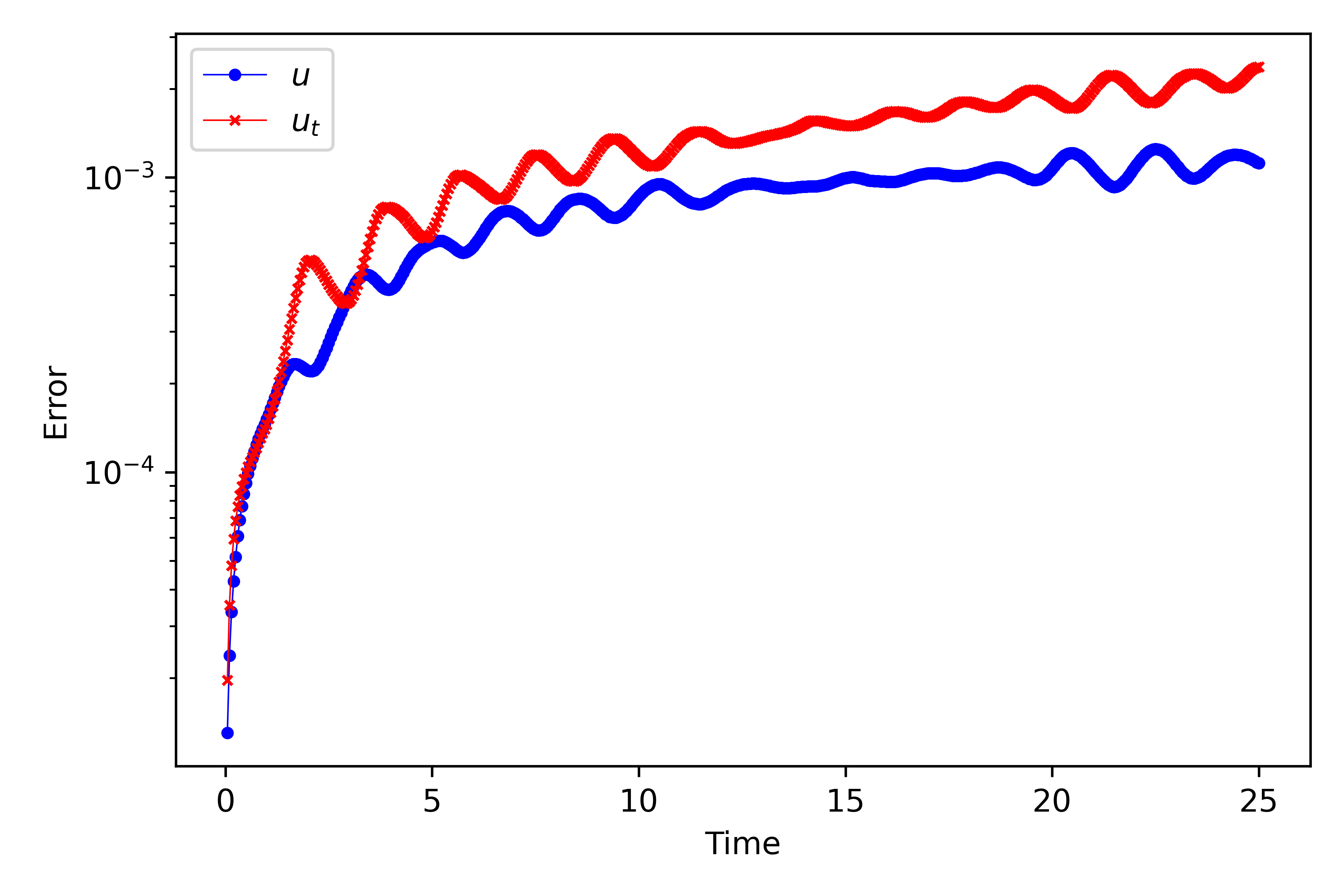}
    (b)\includegraphics[width=0.4\textwidth]{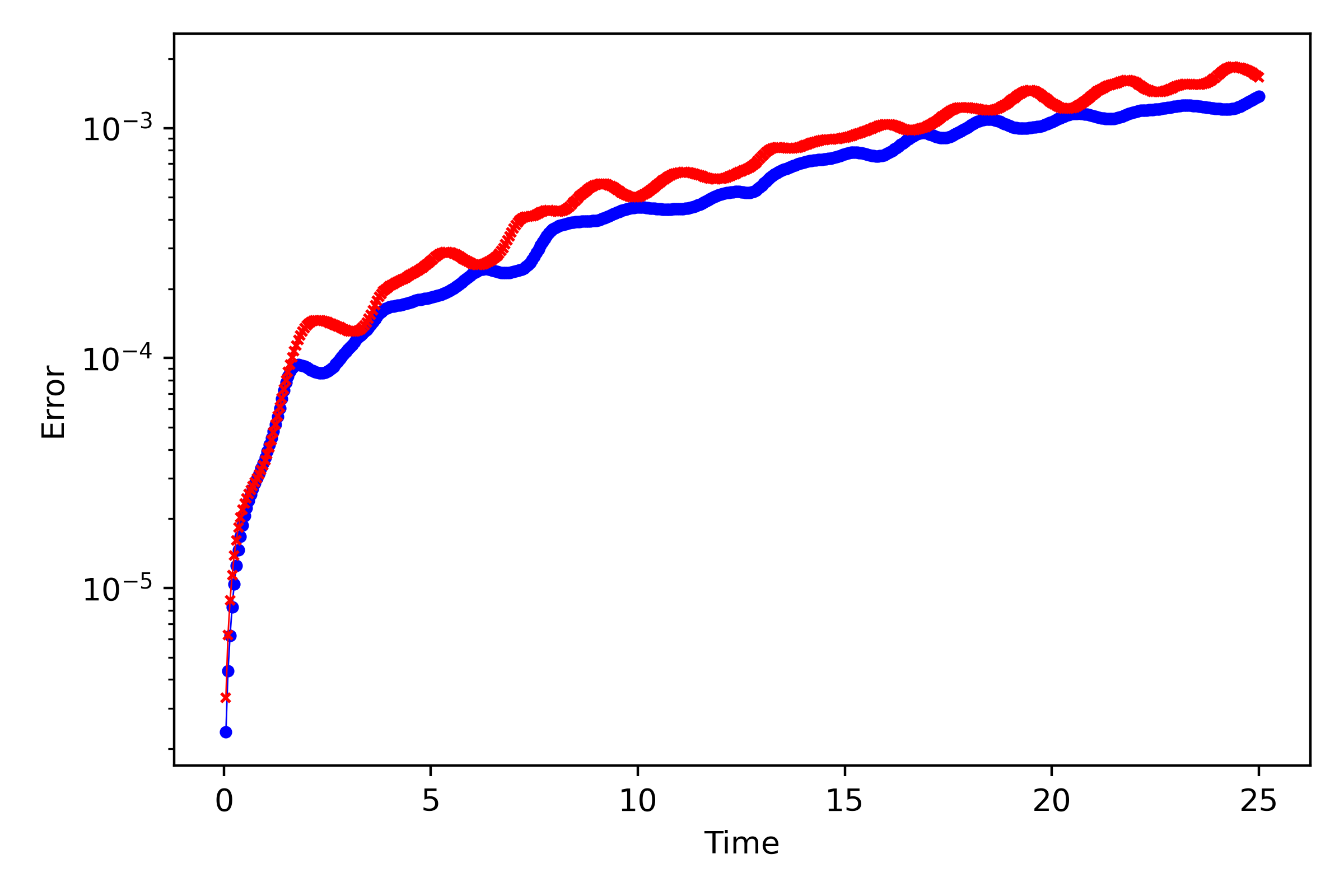}
    \caption{Average relative $\ell_2$ error over 100 test samples from nodal (a) and modal (b) FML.}
	\label{fig:  wave1d error}
\end{figure}

For validation, we independently generate 100 test samples from the training distribution and evaluate
10 independently trained models for 500 time steps (up to $T=25$).
\begin{figure}[htbp]
	\centering
	\includegraphics[width=0.8\textwidth]{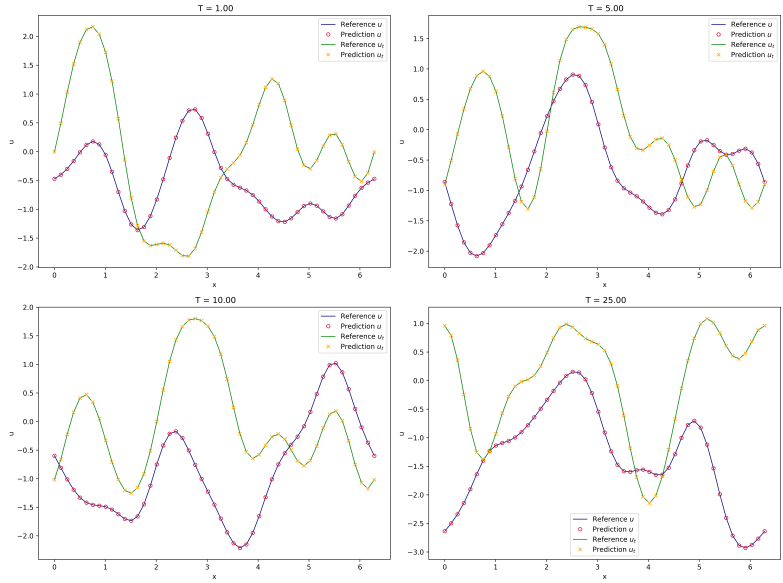}
    \caption{Reference and nodal FML predictions for the solution of \eqref{wave}.}
	\label{fig: wave1d traj}
\end{figure}
\begin{figure}
    \centering
    \includegraphics[width=0.8\textwidth]{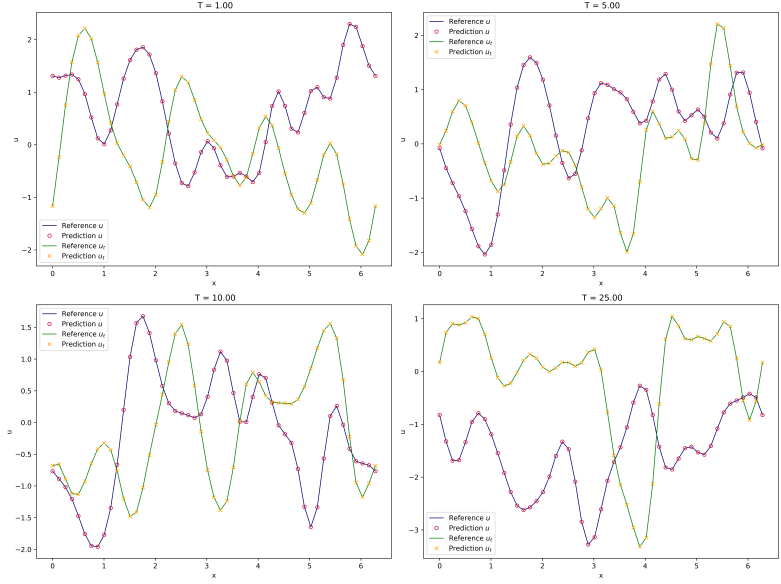}
    \caption{Reference and modal FML predictions 
    for the solution of \eqref{wave}.}
    \label{fig: wave1d modal traj}
\end{figure}
Figures~\ref{fig: wave1d traj} and \ref{fig: wave1d modal traj} show that both methods reproduce
the oscillatory reference trajectory over the 500-step prediction interval. Figure~\ref{fig:  wave1d error}
reports the mean relative discrete $\ell_2$ errors over the 100 test samples.

\subsection{Mixed Nonlocal Diffusion Equation}
Next, we consider the mixed nonlocal diffusion equation with periodic boundary conditions:
\begin{equation}\label{2Ddiffusion}
\left\{
\begin{aligned}
    & \partial_{t} u(x,y,t) = -c_1(-\Delta)^{\frac{\alpha}{2}}u-c_2(-\Delta)^{\frac{\beta}{2}}u,\quad \mbox{for} \ \ (x,y)\in(0,2\pi)^2, \ \ t>0,\\
    & u(x,y,0) = u_0(x,y),\quad \mbox{for} \ \ (x,y)\in[0,2\pi]^2,
\end{aligned}
\right.
\end{equation}
where we choose $\alpha = 1.5$, $\beta = 0.5$, and $c_1 = c_2 = 0.05$. 
The numerical solutions are generated with the Fourier pseudospectral method in space and exact time integration with $N_x=N_y=80$,
using a periodic grid with duplicated endpoints.

We generate $N_{\rm train}=10^6$ training sequences from random initial conditions of the form
\begin{equation}
    u_0(x, y) = \sum_{n=0}^{N_{c}^x}\sum_{m=0}^{N_{c}^{y}}\big(A(n,m)\cos(nx+my)+B(n,m)\sin(nx+my)\big),
\end{equation}
where $N_{c}^x, N_{c}^{y}\sim U\{0,1,2,3,4\}$ and $A(n,m), B(n,m)\sim U[-2^{-(n+m)}, 2^{-(n+m)}]$. 
For each initial condition, we select six consecutive snapshots, corresponding to five recurrent prediction steps:
$$
\left\{\u^{(i)}(t_i),\u^{(i)}(t_i+\Delta t),\ldots,
\u^{(i)}(t_i+5\Delta t)\right\}_{i=1}^{N_{\rm train}},
$$
where $t_i\sim U\{0,\Delta t,\ldots,15\Delta t\}$ and $\Delta t=0.05$.

For nodal FML, we use a disassembly--assembly network with $J=5$, $n_d=n_a=2$, and $n_w=441$.
For modal FML, we use a four-block ResNet with three hidden layers per block and 50 neurons per layer.
Both models use a five-step recurrent loss.

\begin{figure}
    \centering
    \includegraphics[width = 0.6\textwidth]{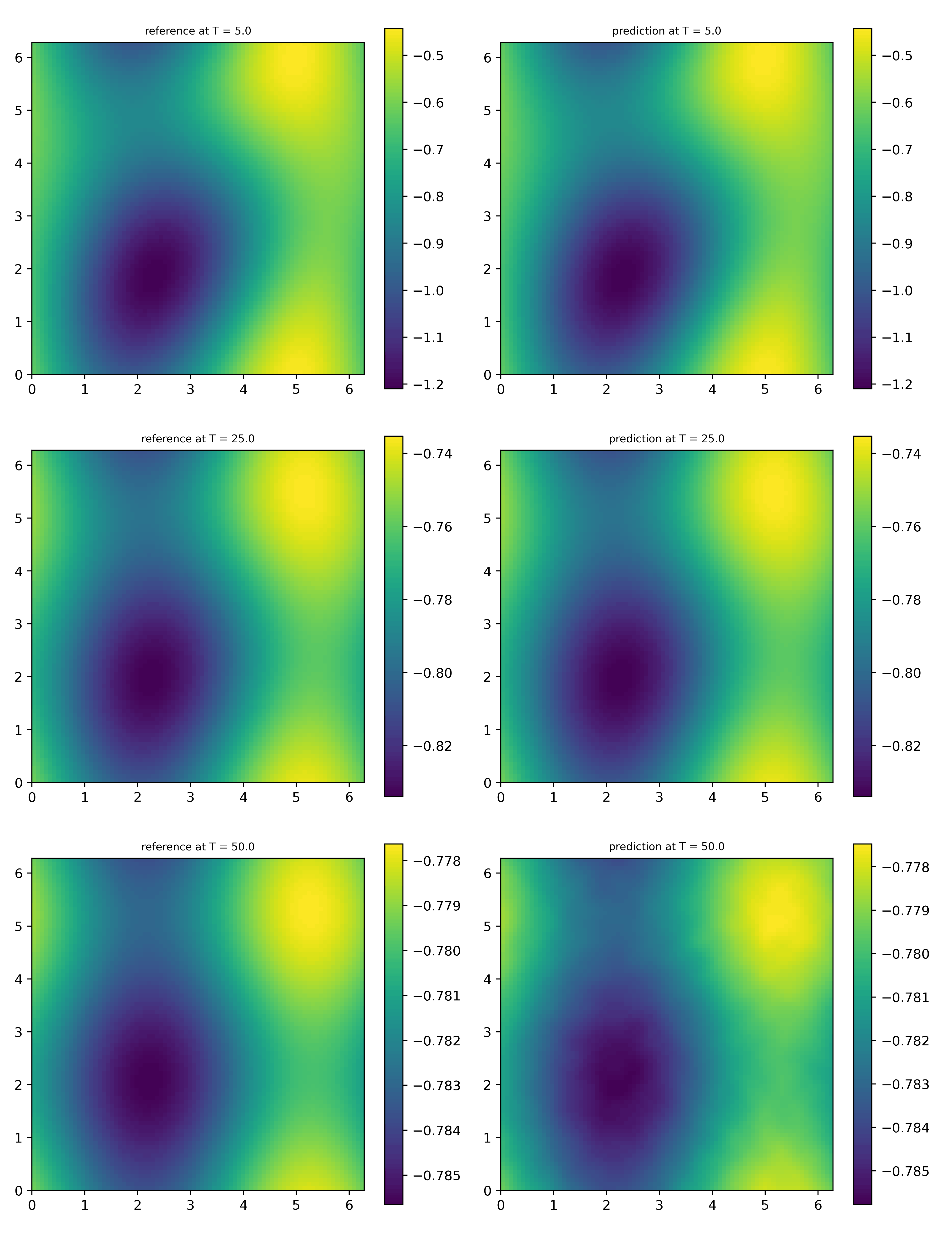}
    \caption{Reference and nodal FML predictions for the solution of \eqref{2Ddiffusion}.}
    \label{fig: diff2d nodal}
\end{figure}

\begin{figure}
    \centering
    \includegraphics[width = 0.6\textwidth]{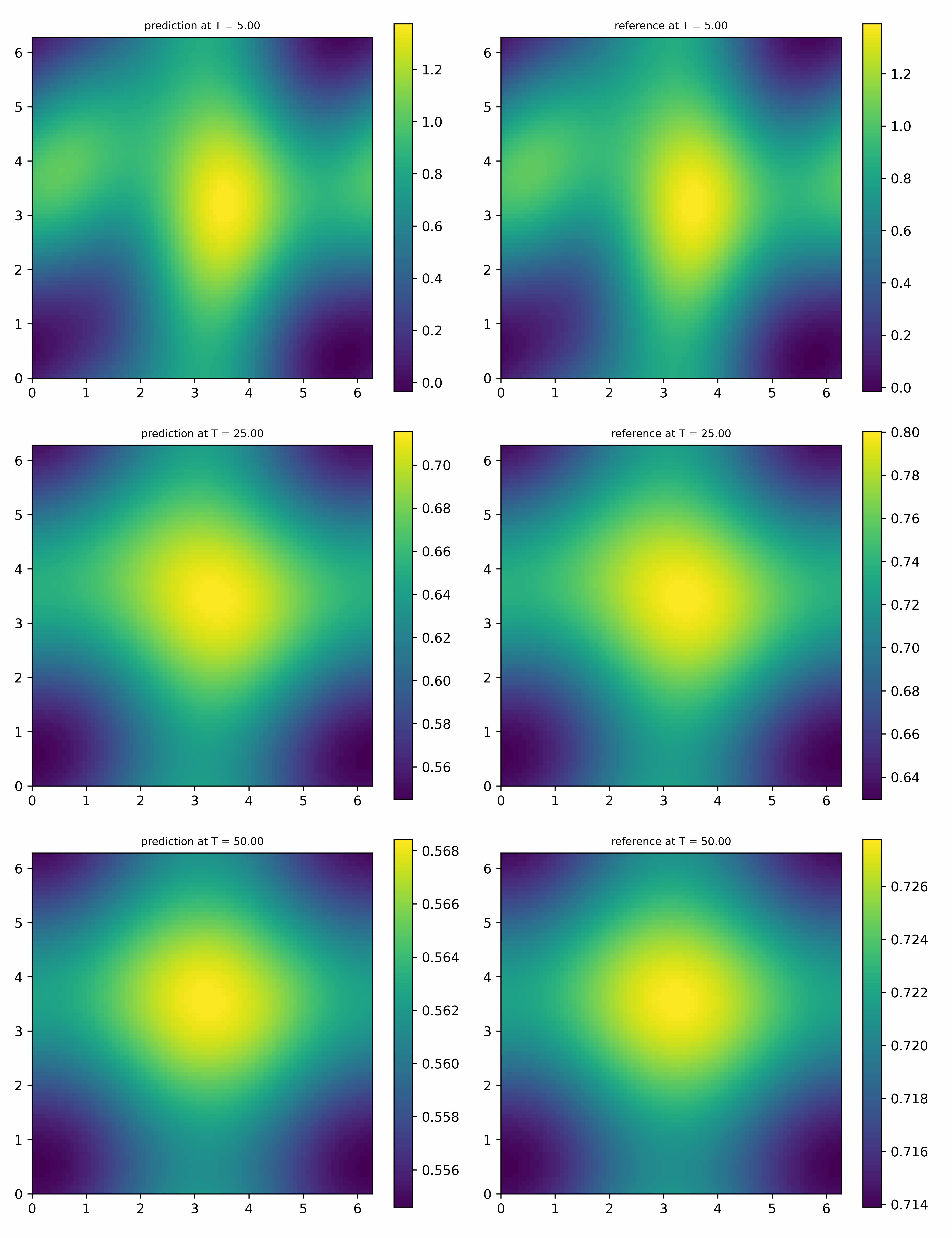}
    \caption{Reference and modal FML predictions for the solution of \eqref{2Ddiffusion}.}
    \label{fig: diff2d modal}
\end{figure}

\begin{figure}[htbp]
	\centering
	\includegraphics[width=0.4\textwidth]{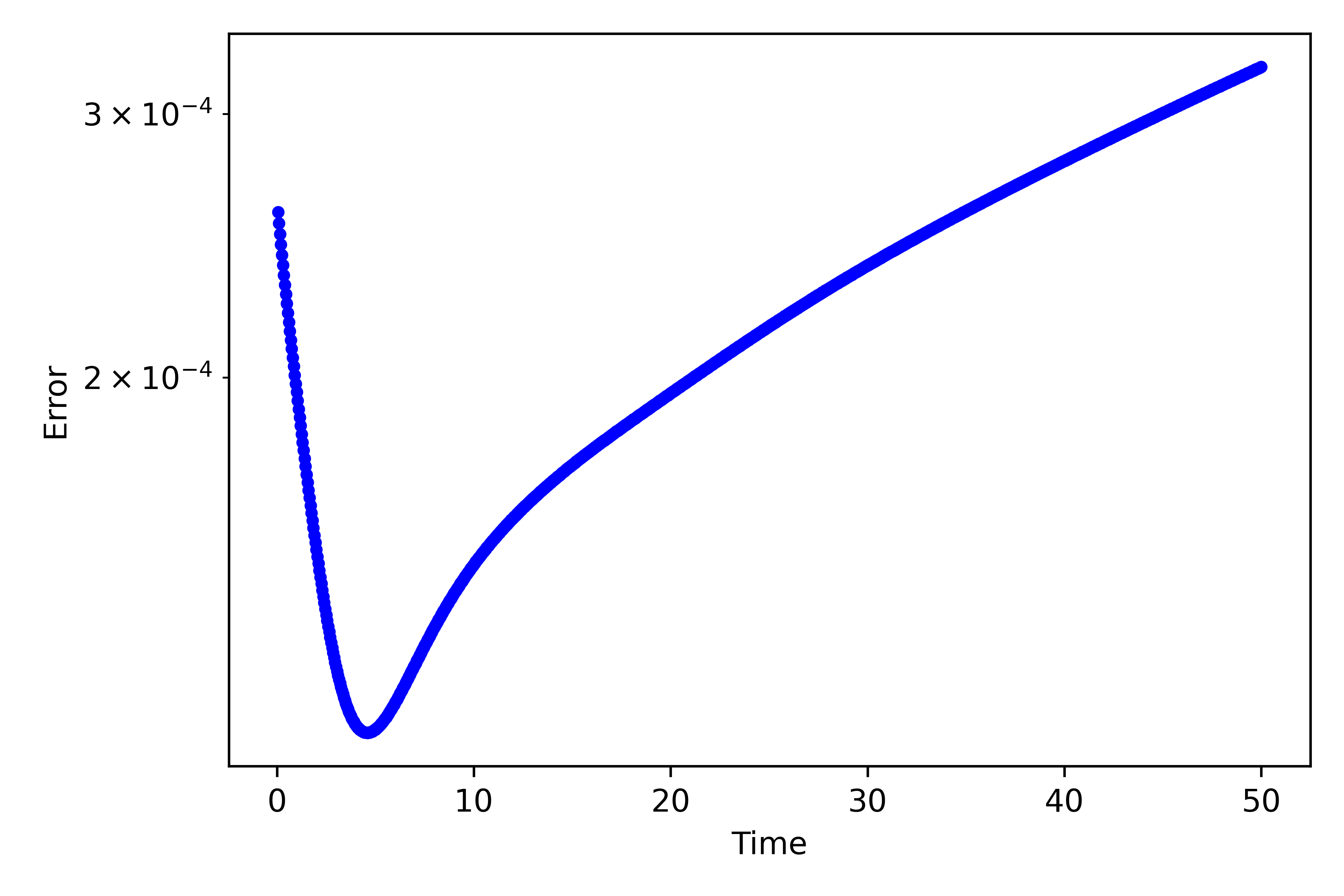}
    \includegraphics[width=0.4\textwidth]{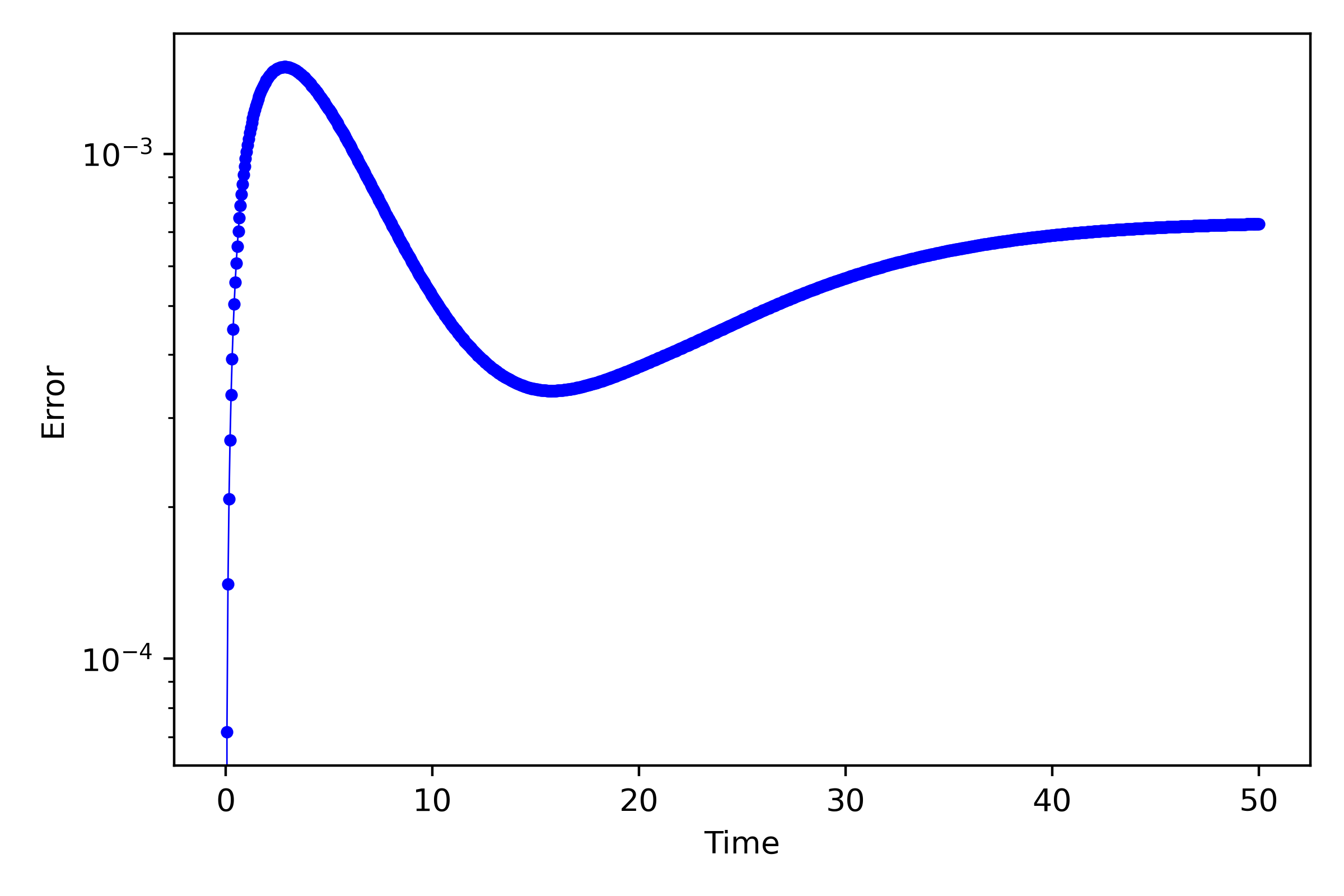}
    \caption{Average relative discrete \texorpdfstring{$\ell_2$}{ell2} error over 100 test samples from nodal
    (left) and modal (right) FML.}
	\label{fig: diff2d error}
\end{figure}

For validation, we independently generate 100 test samples from the training distribution and evaluate
five independently trained models for 1,000 time steps (up to $T=50$). Figures~\ref{fig: diff2d nodal}
and \ref{fig: diff2d modal} compare the reference and predicted solutions for one test sample. Both methods
reproduce the spatial decay pattern throughout the prediction interval. Figure~\ref{fig: diff2d error}
reports the mean relative discrete $\ell_2$ errors over the 100 test samples.

\subsection{Parameter recovery}
We further assess the learned modal FML dynamics by recovering the fractional orders from the predicted
trajectories. To accomplish this, we recall the fractional Laplacian pseudo-differential operator \eqref{eq:fractional_laplacian_fourier},
$$(-\Delta)^{\alpha/2}u(\mathbf{x},t) = \mathcal{F}^{-1}[|\xi|^{\alpha}\mathcal{F}[u]].$$

\begin{figure}
    \centering
        \includegraphics[width = 0.6\textwidth]{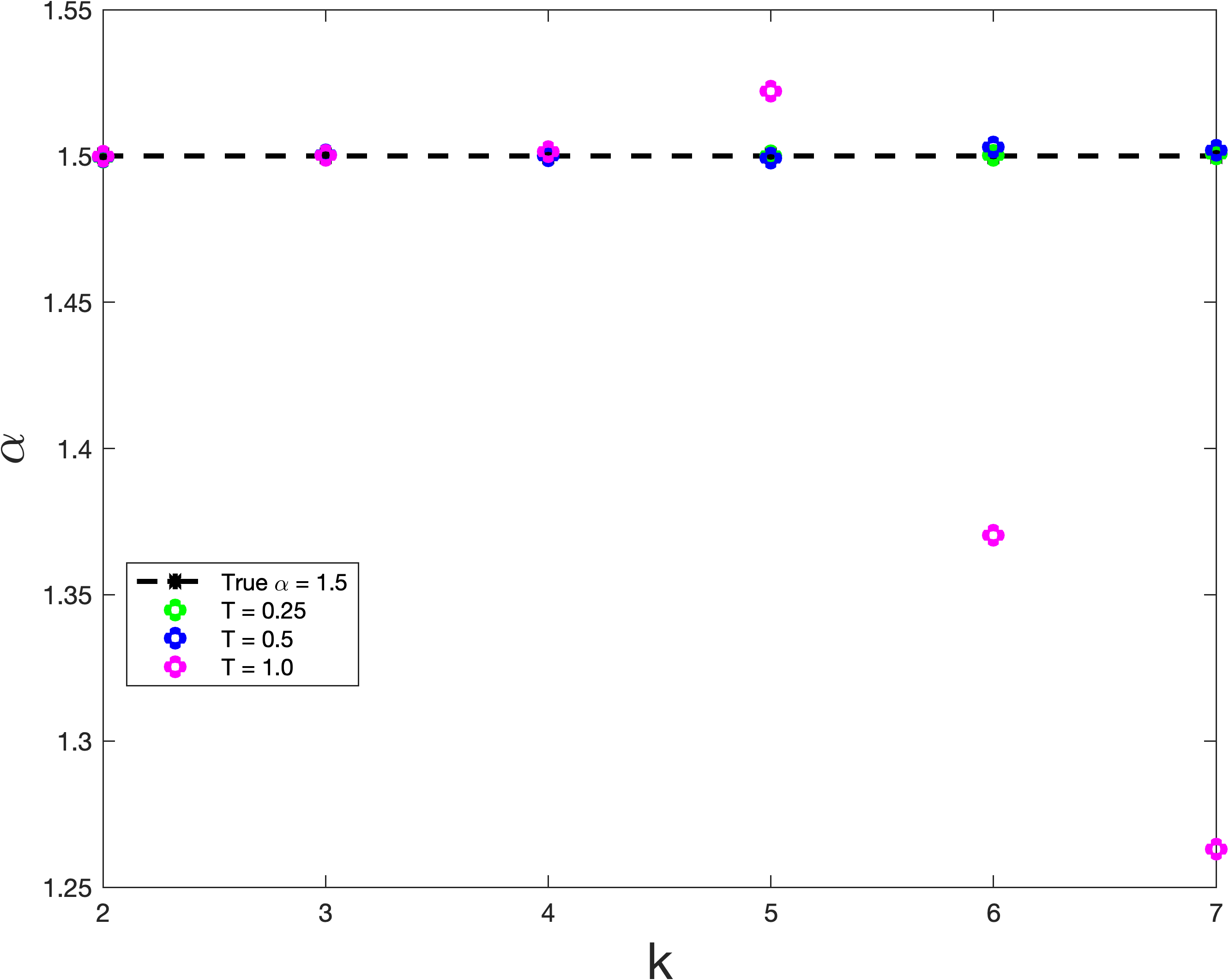}
    \caption{Fractional order $\alpha$ recovered from a predicted trajectory of the 1D diffusion equation,
    plotted against the integer Fourier mode $k$.}
    \label{fig: diffu1D alpha}
\end{figure}

For the 1D diffusion problem, the modal solution satisfies
\begin{equation}
    \hat{u}(\xi,t) = \hat{u}(\xi,0)e^{-|\xi|^{\alpha}t},
\end{equation}
which provides a modewise estimate $\alpha(t,k)$. On the periodic domain, the frequency variable is
the integer Fourier mode $k$, so $\xi=k$. Figure~\ref{fig: diffu1D alpha} shows the estimates from
six nonzero modes of one predicted trajectory over $t\in[0,1]$. The estimates agree closely with the
true value $\alpha=1.5$ at early times. At later times, the modal amplitudes approach zero, making the
estimate increasingly sensitive to numerical and prediction errors.

\begin{figure}
    \centering
    \includegraphics[width = 0.6\textwidth]{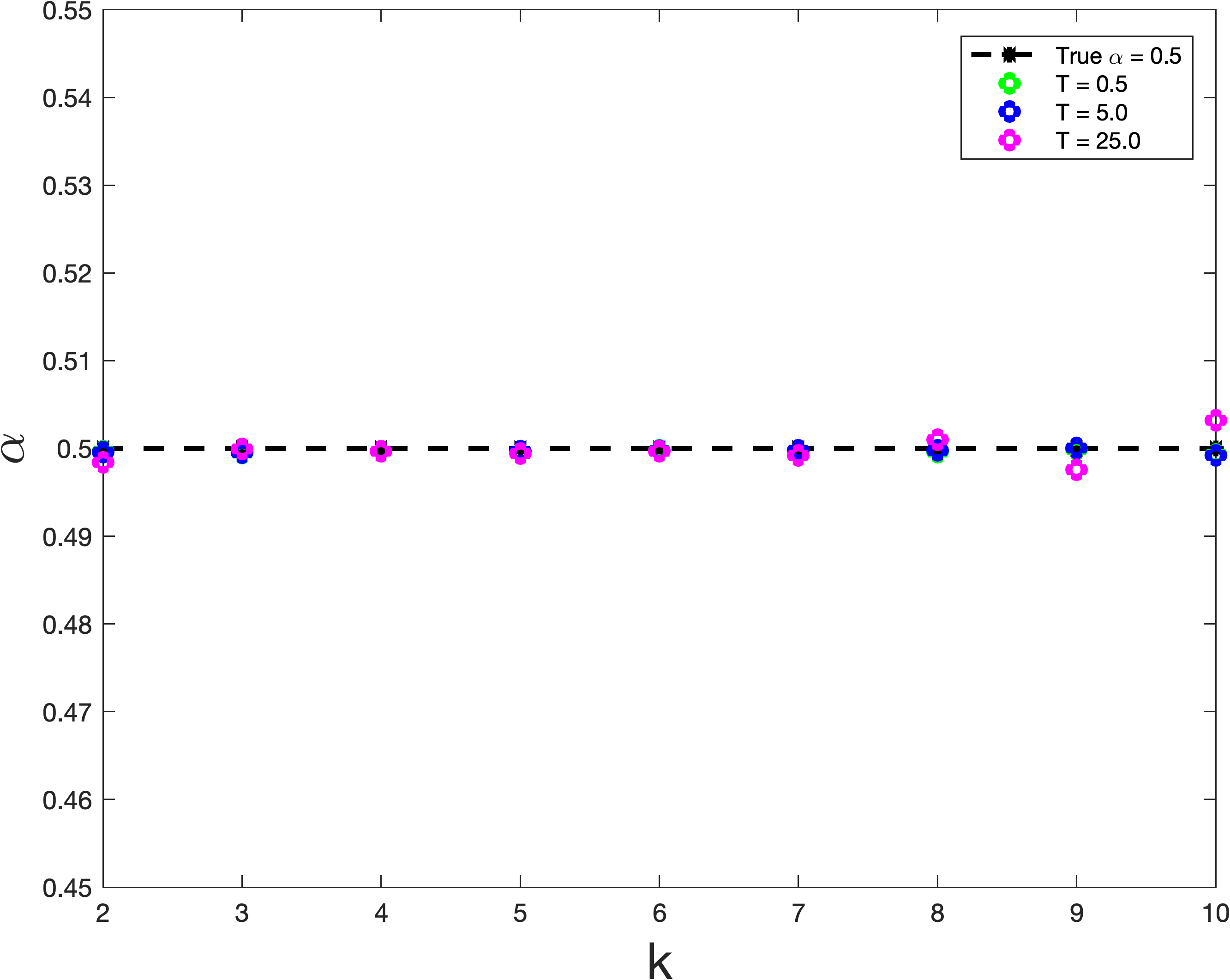}
    \caption{Fractional order $\alpha$ recovered from a predicted trajectory of the 1D nonlocal wave equation.}
    \label{fig: wave1D alpha}
\end{figure}

For the 1D wave equation, the Fourier coefficient $\hat{u}(\xi,t)$ satisfies the ODE
$$\hat{u}_{tt}(\xi,t) +c|\xi|^{\alpha}\hat{u}(\xi,t) = 0.$$
We approximate $\hat{u}_{tt}$ from the predicted modal snapshots using the centered second-order finite
difference associated with the Leap-Frog scheme and use the result to estimate $\alpha$.
Figure~\ref{fig: wave1D alpha} shows that the estimate remains close to the true value over a longer
interval than in the diffusion example. Unlike the diffusive modes, the wave modes do not undergo
monotone amplitude decay, so their signal remains available for parameter recovery at later times.

\begin{figure}
    \centering
    \includegraphics[width = 0.6\textwidth]{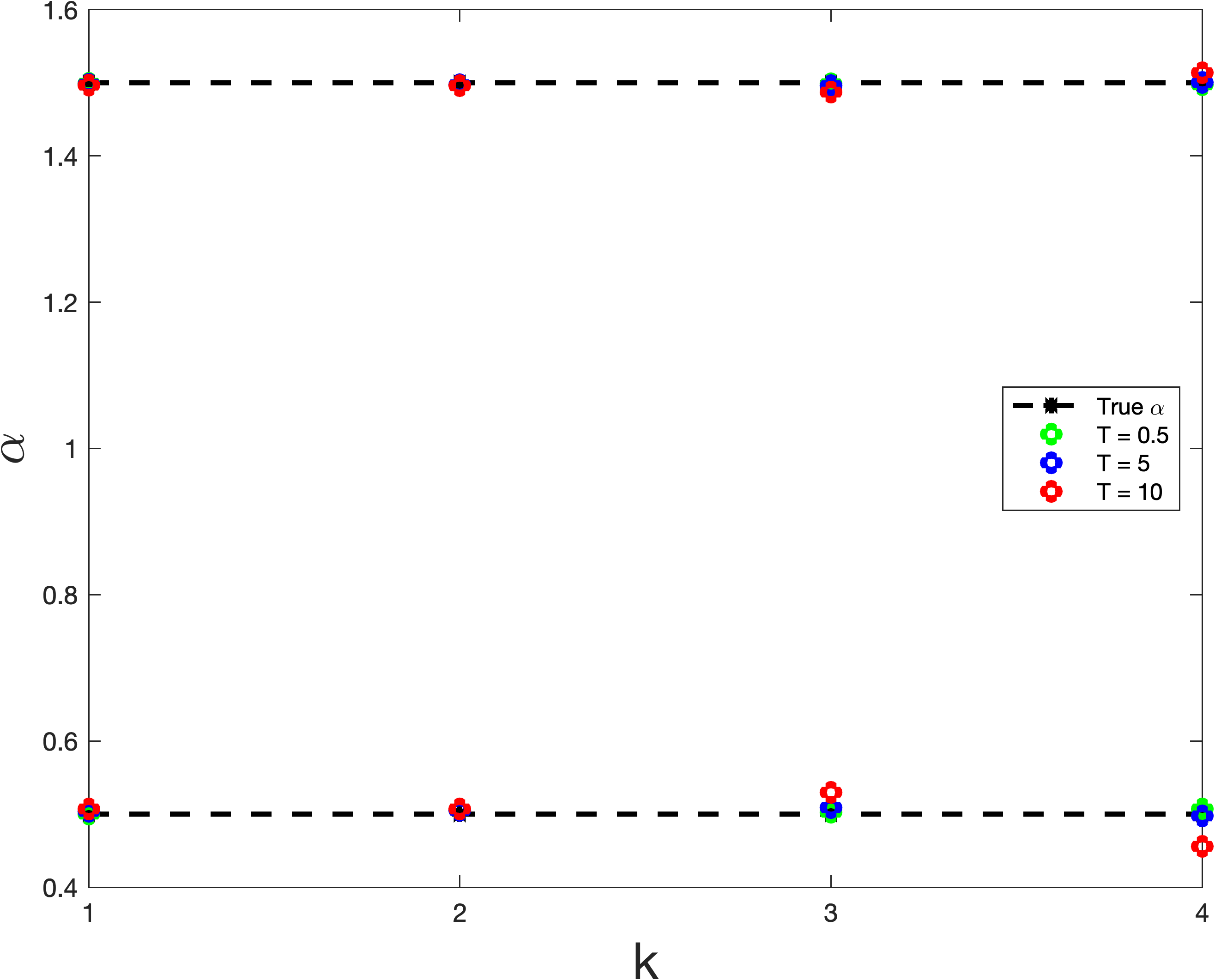}
    \caption{Fractional orders $\alpha$ and $\beta$ recovered from a predicted trajectory of the 2D
    nonlocal diffusion equation.}
    \label{fig: diffu2D alpha}
\end{figure}

For the 2D diffusion equation, $\xi=(k,l)$. We recover $(\alpha,\beta)$ from the predicted modal
coefficients using the MATLAB \lstinline[style=Matlab-editor]{fit} function. For each
$k\in\{1,2,3,4\}$, we fit one parameter pair using the four nonzero modes with
$l\in\{1,2,3,4\}$. Figure~\ref{fig: diffu2D alpha} shows the four resulting estimates.
Because $c_1=c_2=0.05$, the modal amplitudes decay more slowly than in the 1D diffusion example,
and useful estimates can be obtained over a longer time interval.

\section{Conclusion}
\label{sec:conclusion}

We have extended flow map learning (FML) to unknown nonlocal PDE systems in both modal and nodal representations. The proposed formulations learn the finite-time evolution operator directly from solution data without explicit approximation of the underlying nonlocal operators. Numerical experiments demonstrate accurate long-time prediction for several representative fractional diffusion and wave equations. These results suggest that FML provides a promising computational framework for data-driven modeling of unknown nonlocal dynamics. Future work will focus on more general classes of nonlocal operators and applications involving experimentally observed data.

\bibliographystyle{siamplain}
\bibliography{references}
\end{document}